\documentclass[journal,twoside]{IEEEtran}
\usepackage{cite}
\usepackage[pdftex]{graphicx}
\graphicspath{{./}}
\DeclareGraphicsExtensions{.pdf,.png}
\usepackage{amsmath}
\usepackage{amssymb}
\usepackage{mathtools}
\usepackage{xfrac}
\usepackage{nicefrac}
\usepackage{booktabs}

\usepackage{fixltx2e}

\usepackage{url}
\usepackage{color}

\newcommand{\mat}[1]{\mathbf{#1}}

\newcommand{\abs}[1]{{\left|{#1}\right|}}

\newcommand{\R}{\mathbb{R}}
\newcommand{\mydef}{\stackrel{\text{\tiny def}}{=}}

\newcommand{\rshift}[2]{{#1}_{\stackrel{#2\:}{\longrightarrow}}}
\newcommand{\lshift}[2]{{#1}_{\stackrel{\:#2}{\longleftarrow}}}

\newcommand{\T}{\mathsf{T}}

\begin{document}
%
\title{Multiplicative Updates for Convolutional NMF Under $\beta$-Divergence}
%
%
%

\author{Pedro~J.~Villasana~T., Stanislaw~Gorlow,~\IEEEmembership{Member,~IEEE} and Arvind~T.~Hariraman} 
\markboth{}%
{Villasana \MakeLowercase{\textit{et al.}}: $\beta$-CNMF}
%



\maketitle

\begin{abstract}
In this letter, we generalize the convolutional NMF by taking the $\beta$-divergence as the contrast function and present the correct multiplicative updates for its factors in closed form. The new updates unify the $\beta$-NMF and the convolutional NMF. We state why almost all of the existing updates are inexact and approximative w.r.t.\ the convolutional data model. We show that our updates are stable and that their convergence performance is consistent across the most common values of $\beta$. 
\end{abstract}


%
\IEEEpeerreviewmaketitle

\section{Introduction}
%
%
%
%
\IEEEPARstart{N}{onnegative} matrix factorization finds its application in the fields of machine learning and in connection with inverse problems, mostly. It became immensely popular after Lee and Seung derived multiplicative update rules that made the up until then additive steps in the direction of the negative gradient obsolete \cite{Lee1999}. In \cite{Lee2001}, they gave empirical evidence of their convergence to a stationary point, using (a) the squared Euclidean distance, and (b) the generalized Kullback--Leibler divergence as the contrast function. The factorization's origins can be traced back to \cite{Paatero1994,Paatero1997}. 

A convolutional variant of the factorization is introduced in \cite{Smaragdis2004} based on the Kullback--Leibler divergence. The main idea is to exploit temporal dependencies in the neighborhood of a point in the time-frequency plane. In their original form, the updates result in a biased factorization. 
To provide a remedy, multiple coefficient matrices are updated in \cite{Smaragdis2007}, one for each translation, and the final update is by taking the average over all coefficient matrices. 

A nonnegative matrix factor deconvolution in 2D based on the Kullback--Leibler divergence is found in \cite{Schmidt2006}. Not only do the authors give a derivation of the update rules, they show a simple way of making the update rules multiplicative. It may be pointed out that the update rule for the coefficient matrix is different from the one in \cite{Smaragdis2007}. The same guiding principles are applied to derive the convolutional factorization based on the (squared) Euclidean distance in \cite{Wang2009}. But in the attempt to give a formal proof for their update rules, the authors largely reformulate a biased factorization comparable to \cite{Smaragdis2004}. 

In \cite{Cichocki2006}, nonnegative matrix factorization is generalized to a family of $\alpha$-divergences under the constraints of sparsity and smoothness, while the unconstrained $\beta$-divergence is brought into focus in \cite{Fevotte2011}. For both cases multiplicative update rules were given. The properties of the $\beta$-divergence are discussed in detail in \cite{Cichocki2010, Fevotte2011}. The combined $\alpha$-$\beta$-divergence together with the corresponding multiplicative updates can be found in \cite{Cichocki2011}.

In this letter, we provide multiplicative update rules for the factorial deconvolution under the $\beta$-divergence. Furthermore, we argue that the updates in \cite{Smaragdis2004, Smaragdis2007} and in \cite{Schmidt2006} are empirical and/or inexact w.r.t.\ the convolutional data model. According to our simulation, the updates in \cite{Wang2009} do not signify any extra improvement over \cite{Smaragdis2007} despite the additional load. Finally, we show that the exact updates are stable and that their behavior is consistent for $\beta \in {\left\{0, 1, 2\right\}}$.

\section{Nonnegative Matrix Factorization}
\label{sec:nmf}

The nonnegative matrix factorization (NMF) is an umbrella term for a low-rank matrix approximation of the form 
\begin{equation}
\mat{V} \simeq \mat{W} \, \mat{H} 
\label{eq:nmf}
\end{equation}
with $\mat{V} \in \R_{\geqslant 0}^{K \times N}$, $\mat{W} \in \R_{\geqslant 0}^{K \times I}$, and $\mat{H} \in \R_{\geqslant 0}^{I \times N}$, where $I$ is the predetermined rank of the factorization. The letters above help distinguish between visible ($v$) and hidden variables ($h$) that are put in relation through weights ($w$). The factorization is usually formulated as a convex minimization problem with a dedicated cost function $C$ according to 
\begin{equation}
\underset{\mat{W},\,\mat{H}}{\text{minimize}}\ {C{\left(\mat{W}, \mat{H}\right)}} \qquad \text{subject to}\ w_{ki}, h_{in} \geqslant 0 
\end{equation}
with 
\begin{equation}
C{\left(\mat{W}, \mat{H}\right)} = L{\left(\mat{V}, \mat{W}\,\mat{H}\right)} \text{,} 
\label{eq:cost}
\end{equation}
where $L$ is a loss function that assesses the error between $\mat{V}$ and the factorization $\mat{W}\,\mat{H}$.

\subsection{$\beta$-Divergence}

The loss in \eqref{eq:cost} can be expressed by means of a contrast or distance function between the elements of $\mat{V}$ and $\mat{W}\,\mat{H}$. Due to its robustness with respect to outliers for certain values of the input parameter $\beta \in \R$, we resort to the $\beta$-divergence \cite{Basu1998} as a subclass of the Bregman divergence \cite{Bregman1967, Cichocki2010}, which for the points $p$ and $q$ is given by \cite{Cichocki2010} 
\begin{equation}
d_\beta{\left(p, q\right)} = \begin{dcases}
\frac{p^\beta + {\left(\beta - 1\right)} \, q^\beta - \beta \, p \, q^{\beta - 1}}{\beta\,{\left(\beta - 1\right)}} \text{,} & \beta \neq 0, 1 \text{,} \\
p \, \log\frac{p}{q} - p + q \text{,} & \beta = 1 \text{,} \\
\frac{p}{q} - \log\frac{p}{q} - 1 \text{,}  & \beta = 0 \text{.} 
\end{dcases}
\end{equation}
Accordingly, the $\beta$-divergence for matrices $\mat{V}$ and $\mat{W}\,\mat{H}$ can be defined entrywise, as 
\begin{equation}
D_\beta{\left(\mat{V} \parallel \mat{W} \, \mat{H}\right)} \mydef \sum_{k=1}^K\sum_{n=1}^N{d_\beta{\left(v_{kn}, \sum\nolimits_i{w_{ki} \, h_{in}}\right)}}\text{,} 
\label{eq:entrywise_divergence}
\end{equation}
which can further be viewed as the $\beta$-divergence between two (unnormalized) marginal probability mass functions with $k$ as the marginal variable. Note that the $\beta$-divergence has a single global minimum for $\sum_n v_{kn} = \sum_{i, n} w_{ki} \, h_{in}$, $\forall k$, although it is strictly convex only for $\beta \in {\left[1, 2\right]}$ \cite{Cichocki2010, Fevotte2011}.

%

\subsection{Discrete Convolution}

As can be seen explicitly in \eqref{eq:entrywise_divergence}, the weight $w_{ki}$ for the $i$th hidden variable $h_{in}$ at point $\left(k, n\right)$ is applied using the scalar product. Given that $h_i$ evolves with $n$, we can assume that $h_i$ is correlated with its past and future states. We can take this into account by extending the dot product to a convolution in our model. 
Postulating causality and letting the weights have finite support of cardinality $M$, the convolution writes
\begin{equation}
u(n) = {\left(w_{ki} \ast h_i\right)}(n) \mydef \sum_{m = 0}^{M - 1}{w_{ki}(m) \, h_{i, n - m}} \text{.}
\label{eq:conv}
\end{equation}
The operation can be converted to a matrix multiplication by lining up the states $h_{in}$ in a truncated Toeplitz matrix:
\begin{equation}
{\begin{bmatrix}
h_{i, 1} & h_{i, 2} & h_{i, 3} & \cdots & h_{i, N - 1} & h_{i, N} \\
0 & h_{i, 1} & h_{i, 2} & \cdots & h_{i, N - 2} & h_{i, N - 1} \\
\vdots & \vdots & \vdots & \ddots & \vdots & \vdots \\
0 & 0 & 0 & \cdots & h_{i, N - M} & h_{i, N - M + 1}
\end{bmatrix}} \text{.}
\label{eq:toeplitz}
\end{equation}
In accordance with \eqref{eq:nmf}, the convolutional NMF (CNMF) can be formulated as follows to accommodate the structure given in \eqref{eq:toeplitz}, see also \cite{Smaragdis2004, Smaragdis2007}: 
\begin{equation}
\mat{V} \simeq \mat{U} = \sum_{m = 0}^{M - 1}{\mat{W}_m \, \rshift{\mat{H}}{m}} \text{,} 
\label{eq:cnmf}
\end{equation}
where 
$\rshift{\cdot}{m}$
is a columnwise right-shift operation (similar to a logical shift in programming languages) that shifts all the columns of $\mat{H}$ by $m$ positions to the right, and fills the vacant positions with zeros. The operation is size-preserving. It can be seen that the CNMF has $M$ times as many weights as the regular NMF, whereas the number of hidden states is equal.

\section{$\beta$-CNMF}
\label{sec:scnmf}

Ensuing from the preliminary considerations in Section~\ref{sec:nmf}, we adopt the formulation of the CNMF from \eqref{eq:cnmf} and derive multiplicative update rules for gradient descent, while taking the entrywise $\beta$-divergence from \eqref{eq:entrywise_divergence} as the loss function. The result is referred to as the $\beta$-CNMF. 

\subsection{Problem Statement}

Under the premise that $\mat{V}$ is factorizable into $\left\{\mat{W}_m\right\}$ and $\mat{H}$, $m = 0, 1, \dots, M - 1$, and given the cost function 
\begin{equation}
C{\left({\left\{\mat{W}_m\right\}}, \mat{H}\right)} = D_\beta{\left(\mat{V} \parallel \mat{U}\right)} \text{,} 
\end{equation}
we seek to find the multiplicative equivalents of the iterative update rules for gradient descent: 
\begin{subequations}
\begin{align}
\mat{W}_m^{t + 1} &= \mat{W}_m^t - \kappa \, \frac{\partial}{\partial \mat{W}_m^t} \, C{\left({\left\{\mat{W}_m^t\right\}}, \mat{H}^t\right)} \text{,} \label{eq:w} \\ 
\mat{H}^{t + 1} &= \mat{H}^t - \mu \, \frac{\partial}{\partial \mat{H}^t} \, C{\left({\left\{\mat{W}_m^t\right\}}, \mat{H}^t\right)} \text{,} \label{eq:h}
\end{align}
\label{eq:gradient_descent}%
\end{subequations}
where \eqref{eq:w} and \eqref{eq:h} alternate at each iteration ($t \geqslant 0$). The step sizes $\kappa$ and $\mu$ are allowed to change at every iteration.

\subsection{Multiplicative Updates Rules}

Computing the partial derivatives of $C$ w.r.t.\ $\mat{W}_m$ and $\mat{H}$, and by choosing appropriate values for $\kappa$ and $\mu$, the iterative update rules from \eqref{eq:gradient_descent} become multiplicative in the form 
\begin{subequations}
\begin{align}
\mat{W}_m^{t + 1} &= \mat{W}_m^t \circ {\left[{\mat{U}^{t^{\circ{\left(\beta - 1\right)}}} \, \rshift{\mat{H}}{m}^{t^\T}}\right]}^{\circ{-1}} \nonumber \\ 
&\qquad{} \circ {\left[{\mat{V} \circ \mat{U}^{t^{\circ{\left(\beta - 2\right)}}}}\right]} \, \rshift{\mat{H}}{m}^{t^\T} \text{,} \label{eq:mur_wm} \\ 
\mat{H}^{t + 1} &= \mat{H}^t \circ {\left[{\sum\nolimits_m{\mat{W}_m^{t^\T} \, \lshift{\widetilde{\mat{U}}}{m}^{t^{\circ{\left(\beta - 1\right)}}}}}\right]}^{\circ{-1}}  \nonumber \\ 
&\qquad{} \circ \sum\nolimits_m{\mat{W}_m^{t^\T} \, {\left[{\lshift{\mat{V}}{m} \circ \lshift{\widetilde{\mat{U}}}{m}^{t^{\circ{\left(\beta - 2\right)}}}}\right]}} \text{,} \label{eq:mur_h} 
\end{align}
\label{eq:multiplicative}%
\end{subequations}
for $m = 0, 1, \dots, M - 1$, with 
\begin{equation}
\widetilde{\mat{U}}^{t} = \sum\nolimits_m{\mat{W}_m^{t+1} \, \rshift{\mat{H}}{m}^t} \text{,} 
\label{eq:mur_u}
\end{equation}
where $\circ$ denotes the Hadamard, i.e.\ entrywise, product, ${\cdot}^{\circ p}$ is equivalent to entrywise exponentiation and $\cdot^{\circ{-1}}$, respectively, stands for the entrywise inverse. The $\lshift{\cdot}{m}$ operator is the left-shift counterpart of the right-shift operator. The details of the derivation of \eqref{eq:multiplicative} can be found in the Appendix. 

\subsection{Relation to Previous Works}

Several multiplicative updates for the CNMF can be found in the existing literature using different loss functions. In \cite{Smaragdis2004}, the loss function is stated as 
\begin{equation}
L{\left(\mat{V}, \mat{U}\right)} = \sqrt{\sum_{k=1}^K\sum_{n=1}^N{{\abs{v_{kn} \, \log{\frac{v_{kn}}{u_{kn}}} - v_{kn} + u_{kn}}}^2}} 
\label{eq:frobenius_loss}
\end{equation}
and the corresponding update rules for $\mat{W}_m$ and $\mat{H}$ are 
\begin{subequations}
\begin{align}
\mat{W}_m^{t + 1} &= \mat{W}_m^t \circ {\left[{\mat{1} \, \rshift{\mat{H}}{m}^{t^\T}}\right]}^{\circ{-1}} \circ {\left[{\mat{V} \circ \mat{U}^{t^{\circ{-1}}}}\right]} \, \rshift{\mat{H}}{m}^{t^\T} \text{,} \label{eq:mu_1_wm} \\ 
\mat{H}^{t + 1} &= \mat{H}^t \circ {\left[{{\mat{W}_m^{t^\T} \, \mat{1}}}\right]}^{\circ{-1}} \circ {\mat{W}_m^{t^\T} \, {\left[{\lshift{\mat{V}}{m} \circ \lshift{\widetilde{\mat{U}}}{m}^{t^{\circ{-1}}}}\right]}} \text{,} \label{eq:mu_1_h} 
\end{align}
\label{eq:mu_1}%
\end{subequations}
where the $\mat{1}$-matrix is of size $K$-by-$N$ with ${\begin{bmatrix}\mat{1}\end{bmatrix}}_{kn} = 1$. At $t$, for every $m$, $\mat{H}$ is updated first using $\mat{W}_m^{t}$ and subsequently $\mat{W}_m^{t + 1}$ is computed from the updated $\mat{H}$, or vice versa. In \cite{Smaragdis2004} it is mentioned and more explicitly stated in \cite{Smaragdis2007} that to avoid bias in $\mat{H}$ towards $\mat{W}_{M - 1}$ it is best to first update all $\left\{\mat{W}_m\right\}$ using $\mat{H}^t$ and to update $\mat{H}$ according to 
\begin{equation}
\begin{aligned}
\overline{\mat{H}}^{t + 1} &= \frac{1}{M} \sum_{m=0}^{M-1}\mat{H}^t \circ {\left[{{\mat{W}_m^{{t+1}^\T} \, \mat{1}}}\right]}^{\circ{-1}} \\ 
&\qquad{} \circ {\mat{W}_m^{{t+1}^\T} \, {\left[{\lshift{\mat{V}}{m} \circ \lshift{\widetilde{\mat{U}}}{m}^{t^{\circ{-1}}}}\right]}} \text{.} 
\end{aligned}
\label{eq:mu_2_h} 
\end{equation}
Comparing \eqref{eq:mu_1} and \eqref{eq:mu_2_h} with \eqref{eq:multiplicative}, it can be noted that both update rules have the same factors for $\beta = 1$. The respective loss function in this case is the generalized Kullback--Leibler divergence $D_1{\left(\mat{V} \parallel \mat{U}\right)}$ and not \eqref{eq:frobenius_loss}. Moreover, the $\mat{1}$-matrix in \eqref{eq:mu_1_h} is not aligned with the $\mat{U}$-matrix. Given that $\beta = 1$, \eqref{eq:mu_1} and \eqref{eq:multiplicative} are identical for $M = 1$, i.e.\ when the NMF is nonconvolutional. For $M > 1$, \eqref{eq:mu_1} is equal to the updates in \cite{Lee2001} for $D_1{\left(\mat{V} \parallel \mat{W} \, \mat{H}\right)}$ for different $\mat{W}_m$-matrices where the $\mat{H}$-matrix is time-aligned via $m$. Eqs.~\eqref{eq:mu_2_h} and \eqref{eq:mur_h} are also different for $M > 1$ because unlike \eqref{eq:mur_h} \eqref{eq:mu_1_h} is the update derived from a nonconvolutional model. $\overline{\mat{H}}$ in \eqref{eq:mu_2_h} brings out the central tendency of the elements in $\mat{H}$ but does not make the factorization in the original loss convolutional. For all the reasons given above, the updates in \eqref{eq:mu_1_h} and \eqref{eq:mu_2_h} are at best an approximation of the update in \eqref{eq:mur_h} for $\beta = 1$. 

In \cite{Schmidt2006}, multiplicative updates are given for a CNMF in 2D (time and frequency) with the (generalized) Kullback--Leibler divergence and the squared Euclidean distance as the loss or cost function. In the dimension of time, the updates are very much the same as the updates in \eqref{eq:multiplicative} for $\beta = 2$. For $\beta = 1$, there is the minor difference that the $\mat{1}$-matrix is not aligned with the $\mat{U}$-matrix, just like in in \eqref{eq:mu_1_h}. 

Other multiplicative updates for a CNMF with the squared Euclidean distance can be found in \cite{Wang2009}. In essence, they are derived in the exact same manner as the updates in \eqref{eq:mu_1}, but for a different loss function, which is $D_2{\left(\mat{V} \parallel \mat{U}\right)}$. Thus, the updates are equal to the ones in \cite{Lee2001} with $\mat{W} = \mat{W}_m$, $\mat{U}$ as in \eqref{eq:cnmf} and $\mat{H}$ time-aligned. For the same reasons that the updates in \eqref{eq:mu_1_h} and \eqref{eq:mu_2_h} are approximative of the update in \eqref{eq:mur_h} for $\beta = 1$, the updates in \cite{Wang2009} are approximative of the update in \eqref{eq:mur_h} for $\beta = 2$. Beyond, \eqref{eq:mur_u} is updated more efficiently as 
\begin{equation}
\widetilde{\mat{U}}^t = \widetilde{\mat{U}}^t + {\left(\mat{W}_m^{t+1} - \mat{W}_m^{t}\right) \rshift{\mat{H}}{m}^t}
\end{equation}
in between the updates of $\mat{W}_m$ and $\mat{W}_{m+1}$. 

\subsection{Interpretation}

The two update rules given in \eqref{eq:multiplicative} are of significant value because, apart from being exact:  
\begin{itemize}
\item They are multiplicative, and thus, they converge fast and are easy to implement. 
\item Eq.~\eqref{eq:mur_wm} extends the update rule of the $\beta$-NMF for $\mat{W}$ to a set of $M$ weight matrices that are linked through a convolution operation. It also extends the corresponding update rule of the existing convolutional NMFs with the squared Euclidean distance or the generalized Kullback--Leibler divergence to the family of $\beta$-divergences. 
\item Eq.~\eqref{eq:mur_h} is even more important, as it yields a complete update of the hidden states at every iteration taking all $M$ weight matrices into account at once. 
\end{itemize}
The update rule in \eqref{eq:mur_h} can be viewed as an equivalent to a descent in the direction of the Reynolds gradient, which is to take the average over the partial derivatives under the group of time translations (in $m$). The average operator reduces the gradient spread as a function of $\mat{W}_m$ at each iteration $t$ such that the loss function converges to a single point that is most likely. 

The $\beta$-NMF being referred to here is the heuristic $\beta$-NMF derived in \cite{Cichocki2006}. It was shown in \cite{Fevotte2011} to converge faster than a computationally equivalent maximization-minimization (MM) algorithm for $\beta \not\in {\left[1, 2\right]}$ and equally fast in the opposite case. The heuristic updates were proven to converge for $\beta \in {\left[0, 2\right]}$, which is the interval of practical value.



\begin{figure*}[!ht]
\centering
  \includegraphics[width=\textwidth]{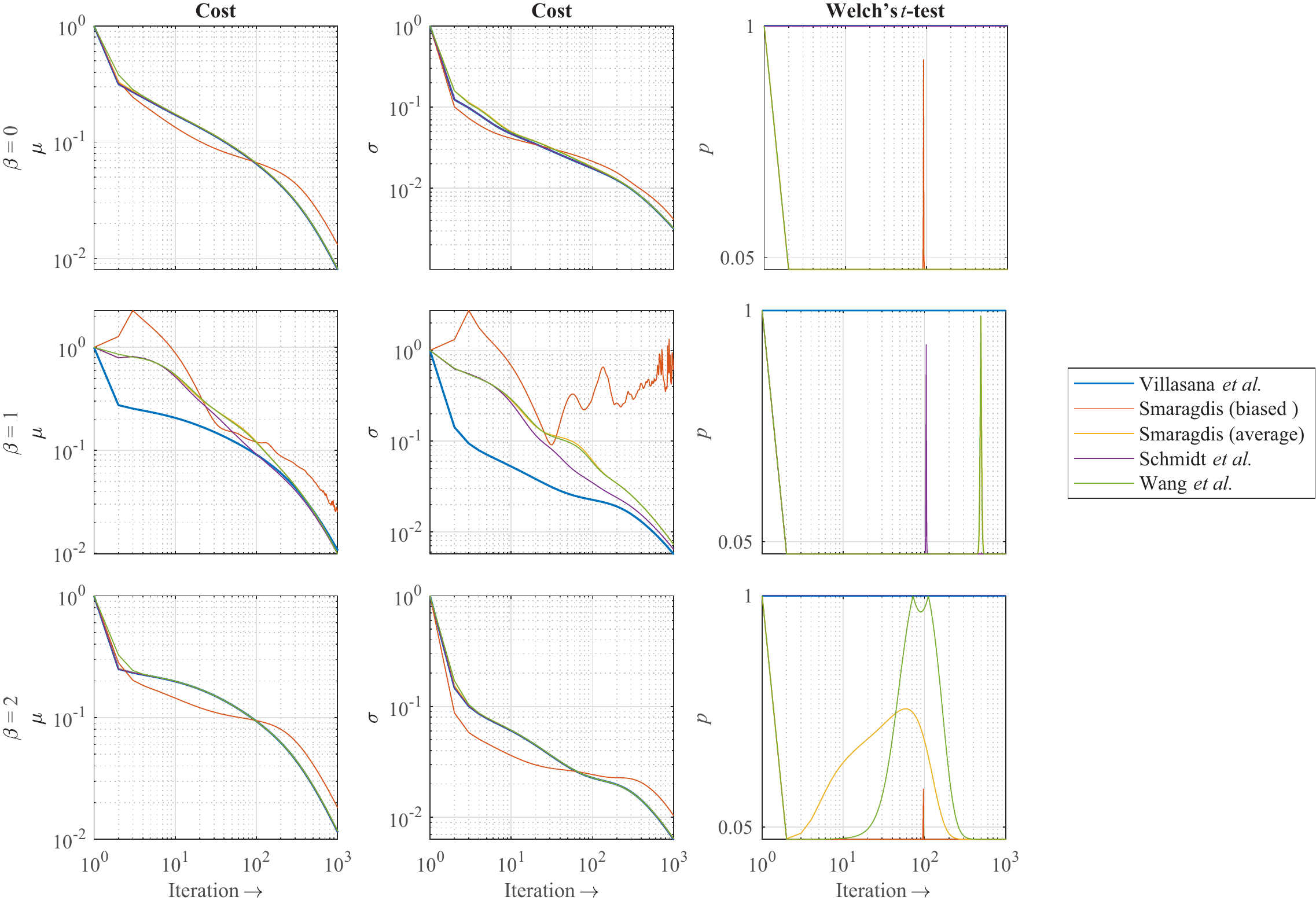}
\caption{Simulation results including the mean and the standard deviation of the loss distribution and the $p$-value from Welch's $t$-test for the hypothesis that the proposed and the existing update rules on average have the same cost (and eventually converge to the same minimum) as a function of iteration.}
\label{fig:simulation}
\end{figure*}

\section{Simulation}

In this section, we compare our proposed updates with the existing ones in terms of convergence behavior and run time for 1000 iterations. To that end, we generate 100 distinct $\mat{V}$-matrices from $M = 16$ $\chi^2$-distributed $\mat{W}_m$-matrices, 
\begin{equation}
w_{ki}(m) = \sum_{p = 1}^2{w_{{ki}_p}^2(m)} \sim \chi^2_2 \qquad w_{{ki}_p}(m) \sim \mathcal{N}{(0, 1)} \text{,} 
\end{equation}
and a uniformly distributed $\mat{H}$-matrix, 
\begin{equation}
h_{in} \sim \mathcal{U}{(0, 1)} \text{.} 
\end{equation}
The factorizations are repeated with 10 random initializations of $\left\{\mat{W}_{m}^{t_0}\right\}$ and $\mat{H}^{t_0}$ with non-zero entries. The results shown in Fig.~\ref{fig:simulation} thus are computed over ensembles of 1000 losses at each iteration. The number of visible and hidden variables is $K = 1000$ and $I = 10$, and the number of realizations (time samples) is $N = 100$. The run time was measured on an Intel Xeon E5-2637 v3 CPU at 3.5 GHz with 16 GB of RAM. In Table~\ref{tab:runtime}, the figures represent the averages over 1000 runs put into relation to the average run time of the proposed updates for 1000 iterations. 

\begin{table}[!t]
\caption{Average Run Time of the Existing Convolutional Updates Relative to the Proposed Updates for Different Betas}
\label{tab:runtime}
\centering
\begin{tabular}{lcccc}
\toprule
& \multicolumn{2}{c}{Smaragdis} & Schmidt \emph{et al.} & Wang \emph{et al.} \\
\cmidrule{2-3}
& Biased & Average \\
\midrule
$\beta = 0$ & 3.42 & 1.05 & 1 & 2.11 \\
$\beta = 1$ & 2.47 & 0.72 & 0.70 & 0.93 \\
$\beta = 2$ & 2.74 & 1.04 & 1 & 1.15 \\
\bottomrule
\end{tabular}
\end{table}

In Fig.~\ref{fig:simulation} it can be seen that the biased updates are clearly least stable under the divergence for which they were meant in the first place ($\beta = 1$). Already in \cite{Schmidt2006}, convergence issues with these updates were reported. For $\beta \in {\left\{0, 2\right\}}$ at less than 100 iterations they can converge faster because $\mat{H}$ and $\mat{U}$ are updated $M$ times per iteration, which explains the significant increase in run time. Between 100 and 1000 iterations, other updates show better performance. Wang's updates are similar in performance to Smaragdis' average updates in spite of the additional intermediate updates of $\mat{U}$ for $\beta = 1$, and slightly worse otherwise. As stated above, Schmidt's updates are the same as ours for $\beta \neq 1$, and so is their behavior. For $\beta = 1$, our updates show the smallest variance overall and yield the the lowest cost below 100 iterations, which is a typical upper bound in practice. The longer run time is due to the shifting of the $\mat{1}$-matrix. For $\beta \in {\left\{0, 2\right\}}$, the loss distributions of our, Wang's, and Smaragdis' average updates look like they have the same mean but a slightly different standard deviation. To test the hypothesis that the costs are statistically equivalent in respect of the mean, we employ Welch's $t$-test. The $p$-values suggest that the null hypothesis can be rejected almost surely for $\beta = 0$, whereas for $\beta = 2$ in general it cannot. 

\section{Conclusion}

To the best of our knowledge, our letter is the only one to provide a complete and exact derivation of the multiplicative updates for the convolutional NMF. Above, the cost function is generalized to the family of $\beta$-divergences. It is shown by simulation that the updates are stable and that their behavior is consistent for $\beta \in {\left\{0, 1, 2\right\}}$. 

\appendix

%

Let $u_{kn} = \sum\nolimits_{i, m}{w_{ki}(m) \, h_{i, n - m}}$ and $\mat{U} = {\begin{bmatrix} u_{kn} \end{bmatrix}} \in \R^{K \times N}$. Then, for any $p \in {\left\{1, 2, \dots, K\right\}}$, $q \in {\left\{1, 2, \dots, I\right\}}$, and $r \in {\left\{0, 1, \dots, M - 1\right\}}$: 
\begin{align}
&\frac{\partial}{\partial {w}_{pq}(r)} \, C{\left({\left\{\mat{W}_m\right\}}, \mat{H}\right)} \nonumber \\
&\qquad{} = \frac{\partial}{\partial {w}_{pq}(r)} \, \sum\nolimits_{n}{\left(\frac{u_{pn}^{\beta}}{\beta} - \frac{v_{pn} \, u_{pn}^{\beta - 1}}{\beta - 1} \right)} \nonumber \\ 
&\qquad{} = \sum\nolimits_n{{\left(u^{\beta - 1}_{pn} - v_{pn} \, u_{pn}^{\beta - 2}\right)} \, h_{q, n - r} } \text{.} 
\end{align}
Choosing $\kappa$ from \eqref{eq:w} as
\begin{equation}
\kappa = \frac{w_{pq}(r)}{\sum\nolimits_n{u_{pn}^{\beta - 1} \, h_{q, n - r}}} \text{,} 
\end{equation}
leads to the first update rule 
\begin{equation}
w_{ki}^{t + 1}(m) = w_{ki}^t(m) \, \frac{\sum\nolimits_n{v_{kn} \, u_{kn}^{t^{\beta - 2}} \, h_{i, n - m}^t}}{\sum\nolimits_n{u_{kn}^{t^{\beta - 1}} \, h_{i, n - m}^t}} \text{.} \qquad \blacksquare
\end{equation}
Further, for any $p \in {\left\{1, 2, \dots, I\right\}}$ and $q \in {\left\{1, 2, \dots, N\right\}}$: 
\begin{align}
&\frac{\partial}{\partial h_{pq}} \, C{\left({\left\{\mat{W}_m\right\}}, \mat{H}\right)} \nonumber \\
&\qquad{} = \frac{\partial}{\partial h_{pq}} \, \sum\nolimits_{k, n}{\left(\frac{u_{kn}^{\beta}}{\beta} - \frac{v_{kn} \, u_{kn}^{\beta - 1}}{\beta - 1}\right)} \text{.}  
\label{eq:partial_C_h}
\end{align}
It is straightforward to show that 
\begin{equation}
\frac{\partial}{\partial h_{pq}} \, u_{kn} = w_{kp}{\left(n - q\right)} 
\label{eq:partial_u_h}
\end{equation}
by setting $n - m = q \leadsto m = n - q$. As a result, plugging in $q + m$ for $n$ in \eqref{eq:partial_C_h} and using \eqref{eq:partial_u_h}, we finally obtain 
\begin{align}
&\frac{\partial}{\partial h_{pq}} \, C{\left({\left\{\mat{W}_m\right\}}, \mat{H}\right)} \nonumber \\
&\qquad{} = \sum\nolimits_{k, m}{w_{kp}(m) \, {\left(u^{\beta - 1}_{k, q + m} - v_{k, q + m} \, u_{k, q + m}^{\beta - 2}\right)}} \text{.} 
\end{align}
Choosing $\mu$ from \eqref{eq:h} as 
\begin{equation}
\mu = \frac{h_{pq}}{\sum\nolimits_{k, m}{w_{kp}(m) \, u_{k, q + m}^{\beta - 1}}} \text{,} 
\end{equation}
leads to the second update rule 
\begin{equation}
h_{in}^{t + 1} = h_{in}^t \, \frac{\sum\nolimits_{k, m}{w_{ki}^t(m) \, v_{k, n + m} \, u_{k, n + m}^{t^{\beta - 2}}}}{\sum\nolimits_{k, m}{w_{ki}^t(m) \, u_{k, n + m}^{t^{\beta - 1}}}} \text{.} \qquad \blacksquare
\end{equation}

\vfill
\clearpage





\bibliographystyle{IEEEtran}
\bibliography{IEEEabrv,references}
\end{document}